\documentclass[letterpaper, 10 pt, conference]{ieeeconf}  
\IEEEoverridecommandlockouts
\usepackage[T1]{fontenc}
\usepackage{amsmath,amssymb,amsfonts}
\usepackage{algorithmic}
\usepackage{graphicx}
\usepackage{textcomp}
\usepackage{xcolor}
\usepackage[export]{adjustbox}
\usepackage{cite}
\def\BibTeX{{\rm B\kern-.05em{\sc i\kern-.025em b}\kern-.08em
		T\kern-.1667em\lower.7ex\hbox{E}\kern-.125emX}}

\usepackage[linesnumbered,lined,boxed,commentsnumbered]{algorithm2e}
\usepackage{multicol}

\usepackage{float}
\usepackage[caption = false]{subfig}
\usepackage{mleftright}

\title{\LARGE \bf
  Accurate position tracking with a single UWB anchor
}

\author{Yanjun~Cao$^{1}$, Chenhao~Yang $^{2}$, Rui~Li$^{3,4}$, Alois Knoll$^{4}$ and
	Giovanni~Beltrame$^{1}$
	\thanks{$^{1}$Y. Cao and Prof.
		Beltrame are with the Department
		of Computer and Software Engineering, \'Ecole Polytechnique de
		Montr\'eal, 2900 Boul \'Edouard-Montpetit, Qu\'ebec
		CA e-mail: (\emph{name.surname}@polymtl.ca).}
	\thanks{$^{2}$C. Yang is with the Wilhelm-Schickard-Institute for
		Computer Science, University of T\"{u}bingen, Germany, e-mail:
		(chenhao.yang@uni-tuebingen.de).}
	\thanks{$^{3,4}$R. Li is with the School of Automation, Chongqing University,
		Chongqing, China and the Department
		of Informatics, Technical University of Munich, Germany, e-mail: (ray.li@tum.de).}
	\thanks{$^{4}$ Prof. Knoll is with the Department
		of Informatics, Technical University of Munich, Germany, e-mail: (knoll@tum.de).}}

\begin{document}

\maketitle
\thispagestyle{empty}
\pagestyle{empty}

\begin{abstract}
  Accurate localization and tracking are a fundamental requirement for robotic
  applications. Localization systems like GPS, optical tracking, simultaneous
  localization and mapping (SLAM) are used for daily life activities,
  research, and commercial applications.  Ultra-wideband (UWB) technology
  provides another venue to accurately locate devices both indoors and
  outdoors. In this paper, we study a localization solution with a single UWB
  anchor, instead of the traditional multi-anchor setup. Besides the challenge
  of a single UWB ranging source, the only other sensor we require is a
  low-cost 9 DoF inertial measurement unit (IMU). Under such a configuration,
  we propose continuous monitoring of UWB range changes to estimate the robot
  speed when moving on a line. Combining speed estimation with orientation
  estimation from the IMU sensor, the system becomes temporally observable. We
  use an Extended Kalman Filter (EKF) to estimate the pose of a
  robot. With our solution, we can effectively correct the accumulated error
  and maintain accurate tracking of a moving robot.
\end{abstract}

\section{Introduction}
Accurate localization and tracking are fundamental services for an autonomous
system. Many options are available for localization: GPS for open outdoor
areas, motion capture systems in a laboratory, visual systems. However, they
are generally limited by the environment or time-consuming and labor-intensive
setup work and expensive
infrastructure\cite{lymberopoulos_microsoft_2015}.  Ultra-wideband (UWB)
technology provides another venue to accurately locate devices both indoors
and outdoors. Most available UWB systems are based on multi-anchor
arrangements, which need some labor-intensive setup work, like mounting
anchors and calibration. Furthermore, it is often difficult to set up such
systems outdoors or in an unstructured environment. We believe that a
localization system which can accurately track devices without complex setup
is highly desirable.


Tracking with a single anchor is attractive because one can easily drop an
anchor in the environment as reference. However, it is also quite challenging:
a single source of distance information is generally too limited for
tracking. Current research in underwater robotics proposes fusing distance to
an acoustic anchor with odometry, but they usually rely on very expensive
sensors (e.g., high accuracy IMU, doppler anemometer)\cite{ferreira_single_2010, reis_source_2018}. Our goal is to enable single anchor localization with low-cost
UWB and IMU sensors. Robots and IoT devices can easily be equipped with these
two sensors, while velocity sensors (encoders, doppler anemometers, etc.) are
much more rare and generally too expensive for IoT devices. Moreover, nowadays
UWB is becoming pervasive, being present in the latest Apple iPhone~\cite{apple} for spatial awareness at the time of writing.

\begin{figure}[tbp]
	\begin{center}
		\includegraphics[trim=0 0 0 1.6cm, width=0.5\textwidth,clip]{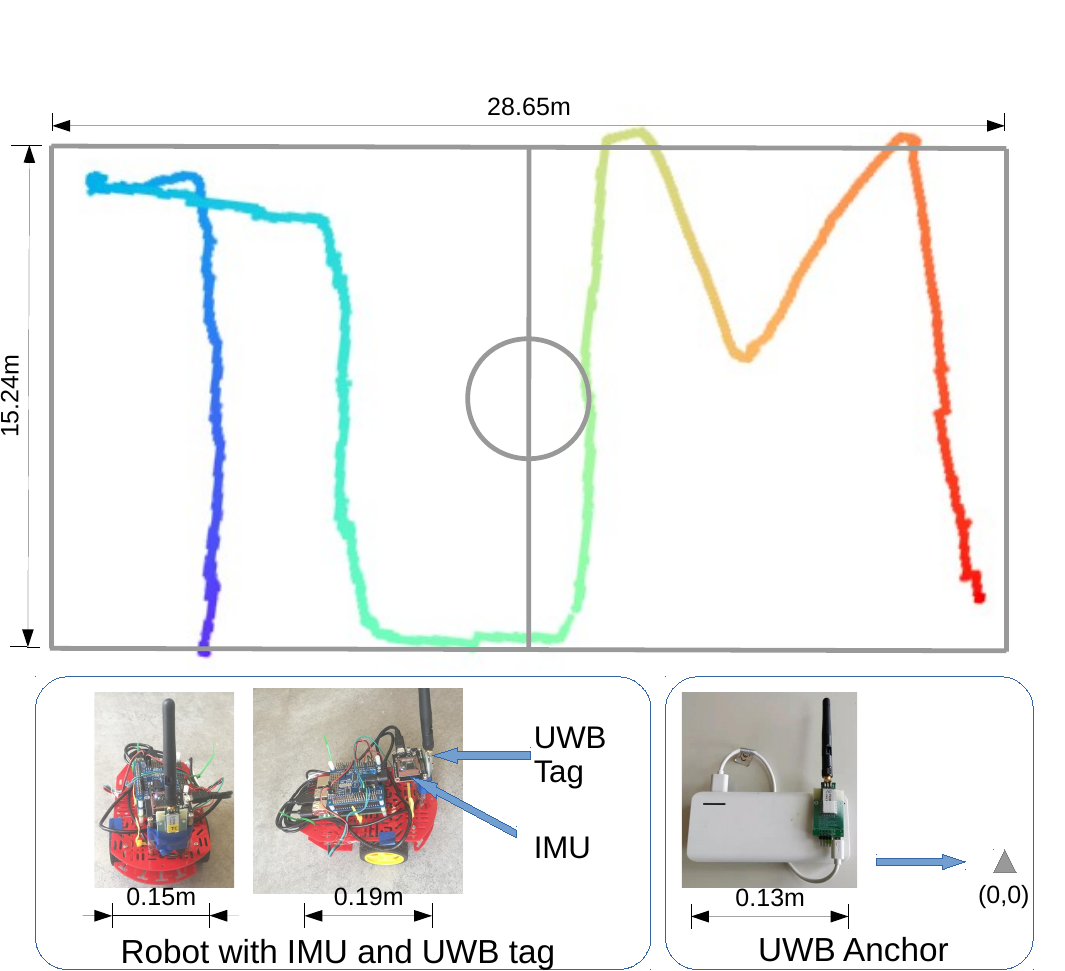}
		\caption{Trajectory of a real-world experiment with a differential wheel robot, Duckiebot. The robot only has IMU and UWB sensors. A UWB anchor is placed on (0,0) in the right bottom. There is no encoder or other velocity sensor used in the robot. The robot is controlled manually to moving a "TUM" like trajectory in a basketball pitch in TUM Garching campus. }
		\label{fig:outdoorExpTraj} 
	\end{center}
\end{figure}

Getting odometry from low-cost IMUs is challenging. Velocity, integrated from
acceleration coming from the IMU, drifts quickly and cannot be used for
odometry. Unfortunately, velocity is crucial for the observability of mobile
robot localization system, especially for single anchor
localization~\cite{martinelli_observability_2005}. To solve this problem, we
propose a novel algorithm to estimate velocity by combining UWB and IMU
measurements. An EKF fuses the range, orientation, and speed estimation to
estimate the robot pose. Simulations and real-world experiments validate our
algorithm.  We believe our system can unlock localization of a large number of
devices in practical applications.

\section{Related Work}

Using UWB technology to locate devices has recently become popular. Most
applications are based on a multi-anchor
configuration~\cite{prorok_accurate_2014 ,miraglia_comparison_2017,
  yao_integrated_2017,gentner_simultaneous_2017,mueller_fusing_2015, nguyen_ultra-wideband-based_2016}. We aim at simplifying the
infrastructure to a single anchor as reference. Many researchers have studied
single anchor localization, especially for underwater
robotics~\cite{ferreira_single_2010,ross2005remarks}. Underwater robots
usually use acoustic sensors, top-of-the-line IMUs, and expensive doppler
sensors. Guo et al.~\cite{guo_ultra-wideband_2017} study a cooperative
relative localization algorithm. They propose an optimization-based
single-beacon localization algorithm to get an initial position for
collaborative localization. However, they only observe a sine-like moving
pattern and they require a velocity sensor. Similar with a recent work proposed by Nguyen et al. ~\cite{nguyen_single_2019}, they also use odometry measurements from optical flow sensors. In our study, we only use UWB and
a low-cost IMU, dropping the need for a velocity sensor.

To better understand the single anchor localization problem, which is
typically non-linear because of distance and angle, an observability study is
necessary. Based on the groundwork of Hermann and
Krener~\cite{hermann_nonlinear_nodate}, researchers have studied the
observability of range-only localization system, from one fixed
anchor~\cite{ross2005remarks} to the relative range and bearing measurements
between two mobile robots~\cite{martinelli_observability_2005}.  Bastista et
al.~\cite{batista_single_2011, indiveri_single_2016} used an augmented state
to linearize the problem, enabling classical observability analysis
methods. A recent study\cite{van_der_helm_board_2019} explores the leader-follower experiment for drones with UWB ranging between robots, also with velocity measurement either from the motion capture system or optical flow.
However, all these studies assume the velocities are available as a
direct measurement, which we do not have.

Although we do not use velocity sensors, the system still needs velocity to
become observable. Getting a reliable velocity from a low-cost IMU or UWB is
challenging: the integration of acceleration drifts dramatically using
low-cost MEMS IMU sensors~\cite{kok_using_2017, woodman_introduction_nodate}.
For position estimation, IMUs are often combined with other sensor
measurements, like GPS, muti-anchor UWB\cite{hol_tightly_2009}, and
cameras~\cite{hol_sensor_2011}.

One straightforward way to estimate velocity is the distance change from a UWB
anchor when the robot is moving along a radical line from the anchor. This
situation is rarely lasting in reality, but the range changing pattern can be
used as a speed estimator.  We propose a method based on simple geometry
relations under the assumption that the robot moves at constant velocity. The
estimated speed coupled with data from the IMU gyroscope can provide a
velocity estimate to keep the system observable. Finally, we use an EKF to
fuse range, orientation, and velocity estimation to get the robot pose. The
contributions for this paper are:




\begin{itemize}
\item a speed estimator using only UWB range information, which changes an
  unobservable system to observable;
\item error analysis for the speed estimator to help design a sensor fusion
  algorithm;
\item a loosely coupled tracking algorithm fusing IMU, UWB, and the proposed
  speed estimation;
\item simulation and real-world experiments to validate our methodology.
\end{itemize}

\section{ Proposed method }

\subsection{System Definition and Observability Analysis}
In this paper, we consider a robot moving on a 2D plane in proximity of a UWB
anchor. The robot has a state vector $\mathbf{x}=[x, y, \theta, v, w]$,
where $x,y$ are the coordinates of the robot, $\theta$ is the heading, and
$v,w$ are linear and angular velocities. The system kinematics are described
as:

\begin{equation}\label{equ:system}
  \dot{\mathbf{x}}=
  \begin{bmatrix}
    \dot{x}\\
    \dot{y}\\
    \dot{\theta}\\
    \dot{v}\\
    \dot{w}\\
  \end{bmatrix}
  =
  \begin{bmatrix}
    v\cos(\theta)\\
    v\sin(\theta)\\
    w\\
    a\\
    b
  \end{bmatrix}
\end{equation}
where $a$ and $b$ are linear and angular acceleration, respectively.  The
measurement functions are
\begin{equation}\label{equ:truemeas}
  \boldsymbol{h(\mathbf{x})} = 
  \begin{bmatrix}
    h_1(\mathbf{x})\\
    h_2(\mathbf{x})
  \end{bmatrix}
  =
  \begin{bmatrix}
    \sqrt{(x-x_A)^2+(y-y_A)^2}\\
    \theta
  \end{bmatrix}
\end{equation}
where $x_A$, $y_A$ are the coordinate of the UWB anchor. $h_1()$ is the range
measurement function for the UWB sensor. $h_2()$ is a heading measurement
function that takes the output orientation of a complementary filter, which
fuses the measurements of the accelerometer, gyroscope, and magnetometer as
the heading measurement of the IMUs.

In control theory, the observability of a system refers to the ability to
reconstruct its initial state from the control inputs and outputs. For a
linear time invariant system, if the observability matrix
$[ C\vert CA \vert ... \vert CA^{n-1} ]$ is nonsingular, the system is
observable \cite{dorf2011modern}.  We refer to an approach by
Hermann~\cite{hermann_nonlinear_nodate} using differential geometry to analyze
the observability of non-linear systems.

To easily compare the impact of velocity, we extend the measurement functions
(\ref{equ:truemeas}) to a typical system that has linear and angular
velocity measurements like (\ref{equ:fullMeas}), similar with \cite{ross2005remarks}. In addition, we define the
anchor position as the origin and map distance measurement $d$ to $\frac{d^2}{2}$ to simplify $h_1(\mathbf{x})$ like in \cite{shi_range_only_2019}. Then we have:

\begin{equation}\label{equ:fullMeas}
\boldsymbol{h(\mathbf{x})} = 
\begin{bmatrix}
	h_1(\mathbf{x})\\
	h_2(\mathbf{x})\\
	h_3(\mathbf{x})\\
	h_4(\mathbf{x})
\end{bmatrix}
=
\begin{bmatrix}
\frac{1}{2}{(x^2+y^2)}\\
\theta\\
v\\
w
\end{bmatrix}
\end{equation}

We rewrite the model in (\ref{equ:system}) to the following format~\cite{hermann_nonlinear_nodate}:
$$\dot{\mathbf{x}} = \sum f^k(\mathbf{x})u_k $$ with a 
state $\mathbf{x} = [x,y, \theta, v, w]$ and control input $u = [a, b]$.
Then we extract a vector field $\boldsymbol{f}$ of the following functions on
the state space:
\begin{align*}
	f^0(\mathbf{x}) &=
	\begin{bmatrix}
	v\cos(\theta), v \sin(\theta), w, 0, 0
	\end{bmatrix}^\top\\
	f^1(\mathbf{x}) &= 
	\begin{bmatrix}
	0,0,0,1,0
	\end{bmatrix}^\top\\
	f^2(\mathbf{x}) &= 
	\begin{bmatrix}
	0,0,0,0,1
	\end{bmatrix}^\top
\end{align*}

Next we find the Lie derivatives of the observation functions on state space
along the vector field $\boldsymbol{f}$.  The zero order Lie derivatives are,
$L^0 h_1 = \frac{1}{2}{(x^2+y^2)}$; $L^0 h_2 = \theta$; $L^0 h_3 = v$; and
$L^0 h_4 = w$, which are same as observation functions.  Then we compute the
first-order derivative as
$L^1_{f^0}h_1 ={xv\cos(\theta)}+{yv\sin(\theta)}{}$; $L^1_{f^0}h_2 = w $;
$L^1_{f^1}h_3 = L^1_{f^2}h_4 = 1$, and all the other first-order Lie derivatives,
$L^1_{f^0}h_3 = L^1_{f^0}h_4 = L^1_{f^1}h_1 = L^1_{f^1}h_2 = L^1_{f^1}h_4 =
L^1_{f^2}h_1 = L^1_{f^2}h_2 = L^1_{f^2}h_3 = 0 $.
	
We write the observation space $\mathcal{G}$ spanned by $L^k_f h_i$, for
$k = 0:1; f = 0:2; i=1:4$. Note that all constant Lie derivatives are
eliminated when computing the state derivative as $d\mathcal{G}$ in
(\ref{equ:orc}). Finally, we compute
the state derivatives of space $\mathcal{G}$ and get:

\begin{equation}\label{equ:orc}
\begin{bmatrix}
d\mathcal{G}
\end{bmatrix}
=
\begin{bmatrix}
o_{11} & o_{12} & 0 & 0 & 0\\
o_{21}& o_{22} &o_{23} & o_{24} & 0\\
0 & 0 & 1 & 0 & 0\\
0 & 0 & 0 & 1 & 0\\
0 & 0 & 0 & 0 & 1
\end{bmatrix}
\end{equation}

where $o_{11} ={x}$,
$o_{12} ={y} $,
$o_{21} =v\cos(\theta) $,
$o_{22} =v\sin(\theta) $,
$o_{23} = -xv\sin(\theta) + yv\cos(\theta)$,
$o_{24} =x\cos(\theta) + y\sin(\theta) $, and
$x,y \neq 0$ because $(0,0)$ is occupied by the anchor.

$d\mathcal{G}$ is with full rank when $v \neq 0$ and $y/x\neq \tan\theta$,
which are reasonable assumptions. If the robot is static, it is difficult to
locate the robot just from the range and the orientation. The second
condition means that the robot moves along a radial line from the anchor,
which is rarely lasting in practice. However, if we do not have velocity
measurements, which is the situation we proposed, the fourth row of the state space
becomes $\boldsymbol{0}_{1\times5}$. The dimension of space is reduced to four,
and therefore the system does not meet the observability rank condition.

Therefore, velocity is crucial for system observability, and we estimate it
from inertial measurements and UWB ranging.



\subsection{Speed Estimation Model}\label{sec:model}

\begin{figure}[bp]
	\begin{center}
		\includegraphics[width=0.4\textwidth,center]{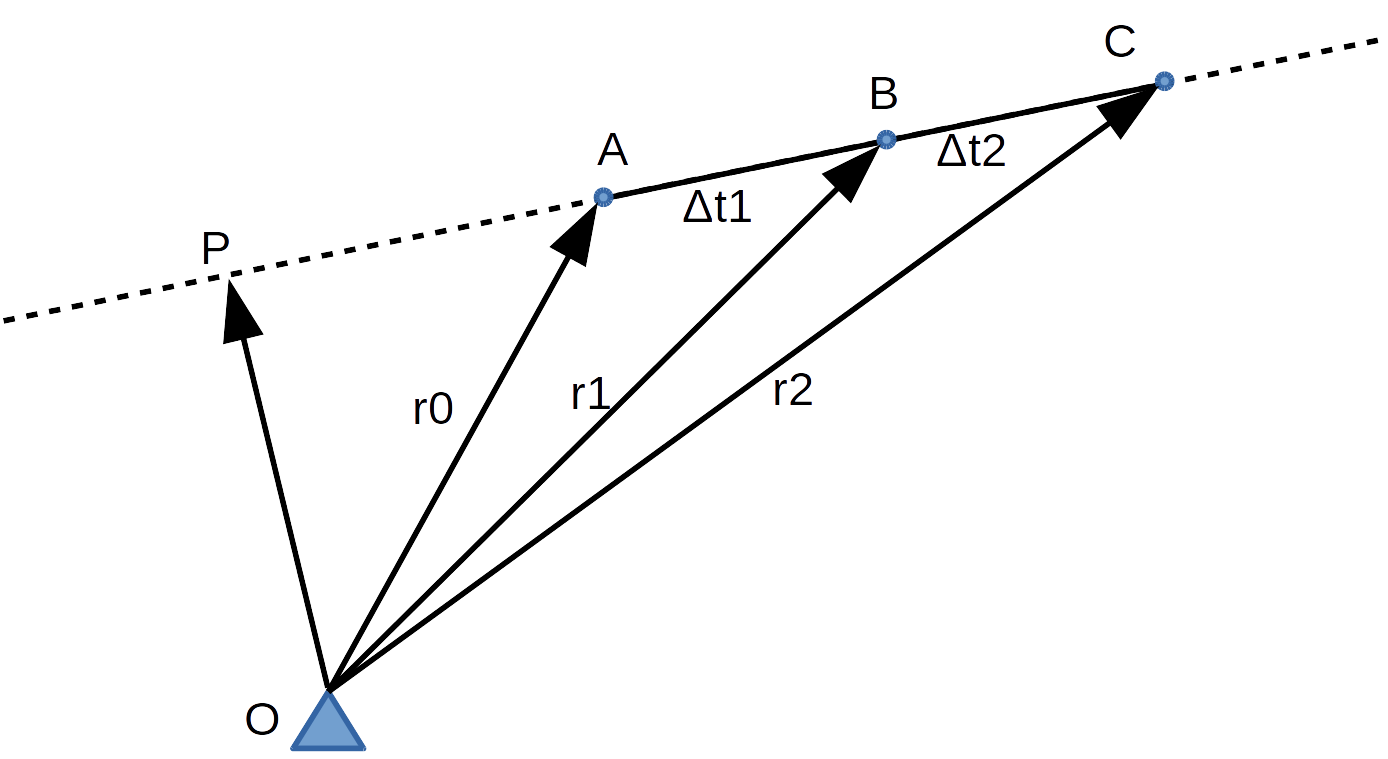}
		\caption{An illustration of our speed estimator. Three pairs of range measurements are used to calculate the speed.}
		\label{fig:velEstimateMath} 
	\end{center}
\end{figure}

By observing the range change pattern as a robot moves on a straight line, we
propose a speed estimator based on simple geometric relations. As shown in
Fig.\ref{fig:velEstimateMath}, three pairs of range and time measurements,
$(r_0, t_0)$ , $(r_1, t_1)$ and $(r_2, t_2)$ are given when the robot passes
points A, B and C with a constant velocity. $r=OP$ is the virtual
distance from the anchor to the motion line. $x=PA$ is the length from
starting point $A$ to the virtual intersection $P$. Based on the Pythagorean
theorem, we can get the algebraic solution of the moving speed $v$:

\begin{equation}
\begin{bmatrix}
{r_0}^2\\
{r_1}^2\\ 
{r_2}^2\\
\end{bmatrix}
 =
\begin{bmatrix}
r^2+x^2\\
r^2+(x+v\Delta t_1)^2\\
r^2+(x+v\Delta t_1+v\Delta t_2)^2\\
\end{bmatrix}
\end{equation}
where $$\Delta t_1 = t_1 - t_0;\Delta t_2 = t_2 - t_1$$
From these three functions, we can solve $r,x$ and $v$.
As we are interested only in the velocity, we just show:
\[v = \pm\sqrt{\dfrac{(r^2_2 - r^2_1) - \dfrac{\Delta t_2}{\Delta t_1}\cdot(r^2_1 - r^2_0)}{\Delta t_1\Delta t_2 +\Delta t_1^2}}.\]
For simplicity, assuming the ranging  measurements  have a fixed frequency $\mathsf{f}$, we have $\Delta t_1 = \Delta t_2 = \Delta t = 1/\mathsf{f}$. Then 
\begin{equation}
v = \pm\sqrt{\dfrac{r^2_2 + r^2_0 - 2r^2_1}{2\Delta t^2}} \label{eq:bf_vel}
\end{equation}

\begin{figure}[bp]
	\begin{center}
		\includegraphics[width=0.5\textwidth,center]{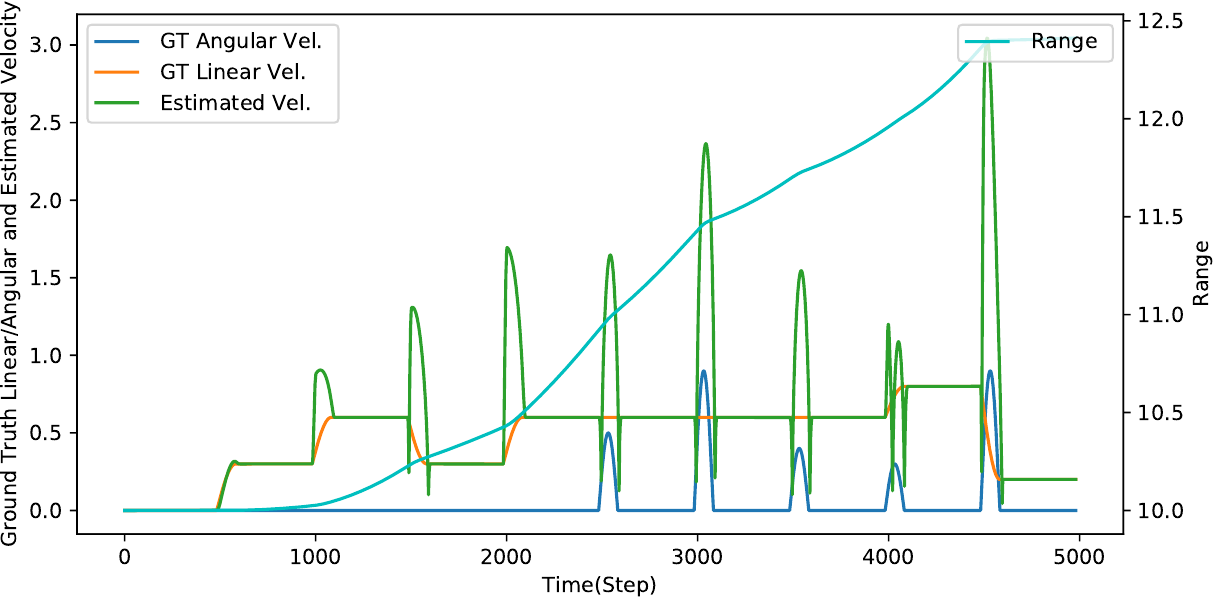}
		\caption{A simulation of speed estimation using noise-free
			ranging measurements to one anchor. The speed can be tracked
			correctly during constant velocity phases. Speed estimation
			produces erroneous peaks during speed change time, which can
			be avoided by filtering.}
		\label{fig:bruteEstimatedVel} 
	\end{center}
\end{figure}

The positive value is the current speed (as the kinematics model shows the
robot can only move forward). We present a simulation that includes ten
stages, with different configurations of linear and angular velocities, as
shown in Fig.\ref{fig:bruteEstimatedVel}. The velocities
change at the beginning of each stage and remain constant thereafter. The range
measurements (cyan) is continuously fed into the estimator. As we can see, the
velocity can be correctly provided (green). The changing pattern of the range
in each phase suggests the speed value.

The peaks between stages are caused by velocity changes, and should be
expected as the changes break the constant velocity assumption. Even if we
cannot estimate the velocity correctly when it is suddenly changing, our
algorithm does not lose its generality, as constant velocity motion is still
the predominant motion in most real-world scenarios. One can estimate the
speed over these periods and maintain a correct pose estimate. Furthermore,
our sensor fusion algorithm also provides tolerance to velocity changes.

\subsection{Speed Estimator Error Analysis} \label{sec:errorAna}

UWB ranging is considered as fairly accurate. This is true compared with WIFI
or Bluetooth technologies that provide meter accuracy, but UWB can only achieve
decimeter accuracy. For instance, DW1000 from Decawave \cite{noauthor_dw1000_nodate} provides the accuracy of $\pm10$ cm using two-way ranging (TWR) time-of-flight (TOF) protocol, which is still too noisy to calculate the speed from 
range measurements directly. In this section, we analyze the error propagation
for the speed estimator and design the speed estimation algorithm accordingly.

The range measurement model is expressed as $r = R + e$, where the measurement $r$ is the true
range $R$ plus some noise $e$. We represent standard deviation as
$\delta *$, e.g., $\delta r$ for the standard deviation of $r$.

To determine the properties of $\delta v$ from three range measurements with deviation $\delta r$, we compute the error propagation
as in~\cite{bohm_introduction_2010} (Ch4). For (\ref{eq:bf_vel}). We get:
\small	
\begin{equation}\label{eq:error1}
	(\delta v)^2 = \left(\dfrac{\partial v}{\partial r_0}\right)^2 (\delta r_0)^2 + \left(\dfrac{\partial v}{\partial r_1}\right)^2(\delta r_1)^2 + \left(\dfrac{\partial v}{\partial r_2}\right)^2(\delta r_2)^2 \nonumber
\end{equation}
\normalsize
this can be rewritten as:

\small	
\begin{align*}
	(\delta v)^2 =& \left(\frac{r_0} {\sqrt{2}\Delta t\sqrt{r^2_2 + r^2_0 - 2r^2_1}}\right)^2 (\delta r_0)^2 \\
	& + \left(\dfrac{r_2} {\sqrt{2}\Delta t\sqrt{r^2_2 + r^2_0 - 2r^2_1}}\right)^2 (\delta r_2)^2 \\
	& +\left(\dfrac{-\sqrt{2}r_1} {\Delta t\sqrt{r^2_2 + r^2_0 - 2r^2_1}}\right)^2  (\delta r_1)^2
\end{align*}
\normalsize

\begin{figure}[tbp]
	\begin{center}
		\includegraphics[width=0.48\textwidth,center]{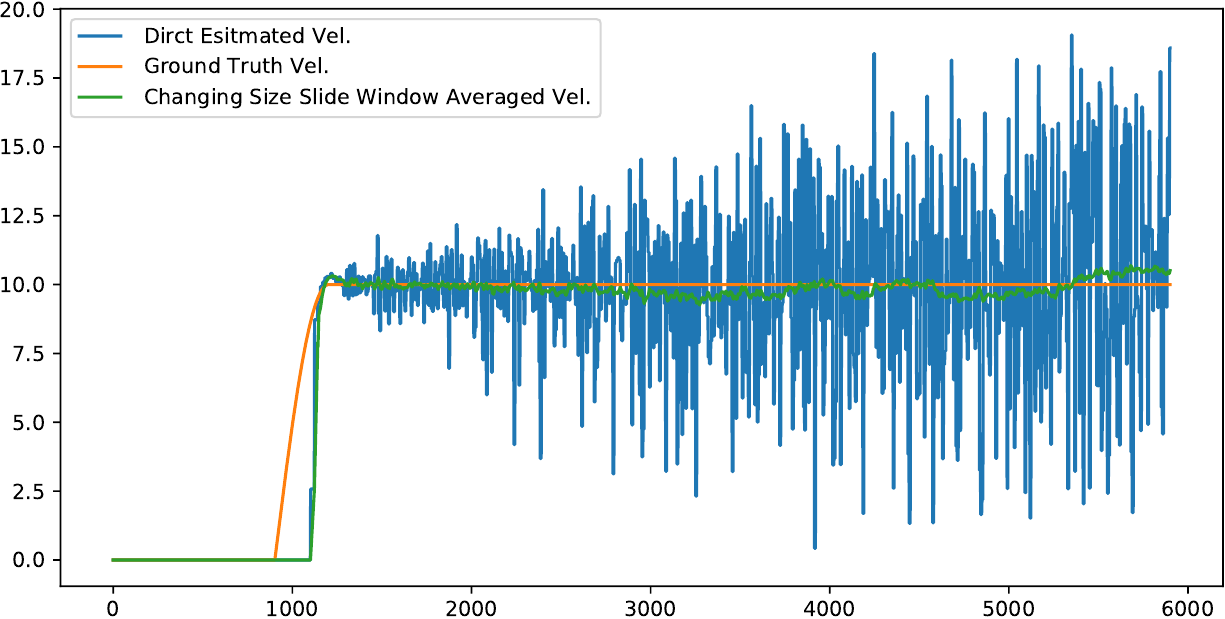}
		\caption{The velocity amplitude is estimated only from range
			measurements. The robot moves from the point (10,0) to (10,
			250) at 10.0 m/s with a step of 0.01s. 
			The anchor is located at (0,0). The
			deviation of the direct speed estimation (blue) increases
			with the range. Using a variable window size average filter,
			our speed estimator (green) gives the correct speed.}
		\label{fig:velEst2} 
	\end{center}
\end{figure}

Since $\delta r_0$, $\delta r_1$, and $\delta r_2$ are measurements from the
same sensor, we have $\delta r_0 = \delta r_1 = \delta r_2 = \delta r $, which
is $0.2$ for UWB sensors in our situation. $\Delta t$ is a constant variable
depending on the ranging update rate. Then we simplify
the equation above as:

\begin{equation}
	(\delta v)^2 = \dfrac{(2r^2_0+2r^2_2+8r^2_1)\Delta t} {{r^2_2 + r^2_0 - 2r^2_1}}^2 (\delta r)^2\nonumber
\end{equation} 
\noindent 
namely,
\begin{equation}
	(\delta v)^2 = \left(2+\dfrac{10r^2_1} {{r^2_2 + r^2_0 - 2r^2_1}}\right) (\delta r)^2\Delta t^2 \label{eq:errProp}
\end{equation} 
From above equation, we can see if three measurements, $r_0$, $r_1$, and
$r_2$, are similar, the denominator $({r^2_2 + r^2_0 - 2r^2_1})$ becomes
small, and then the error becomes very large, a condition we want to avoid.
Similar $r_0$, $r_1$, and $r_2$ can be caused by an excessively fast sample
rate or by very slow motion.

To avoid the arrival of similar ranging measurements, we update the speed
based on the ranging change, instead of updating at regular intervals as
usual. In other words, when the range measurement difference exceeds a given
threshold, the estimator is triggered to compute the speed. This way, the
noise in the estimation can be effectively eliminated. Note that the update
rate depends on the speed of the robot and also the direction of motion. If
the robot moves quickly and on a radial line from the anchor, the range
measurement difference quickly reaches the threshold and the update rate is
high, and vice versa.

Furthermore, we can also see the standard deviation of the speed is positively
correlated with $r_1$ in the numerator in (\ref{eq:errProp}). The value of
$r_1$ is the intermediate measurement of the distance from the robot to the
anchor. As Fig.\ref{fig:velEst2} shows, the deviation of the speed estimation
(blue) increases as the range measurement increases. Thus, if the robot is
very far away from the anchor, the standard deviation becomes large, leading
to noisy speed estimation.

To solve this problem, we implement a variable window size filter, based on
the distance from the robot to the anchor, to smooth the speed estimation. As
Fig.\ref{fig:velEst2} shows, the deviation of direct speed estimation
increases as the range increases. However, our variable window speed
estimation can track the actual speed well.

\subsection{Sensor Fusion System}

\begin{figure}[tp]
	
	\includegraphics[width=0.48\textwidth,center]{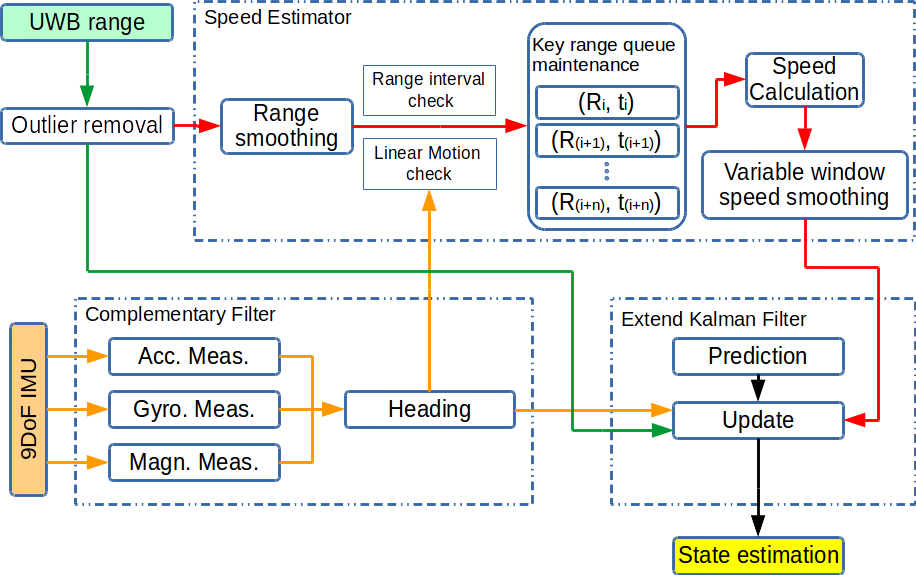}
	\caption{Sensor Fusion system.}
	\label{fig:system} 
	
\end{figure}

The block diagram in Fig.\ref{fig:system} shows our localization system. As
mentioned above, the system has two kinds of sensors: UWB sensors for range
measurement and low-cost IMUs, with gyroscopes, accelerators, and
magnetometers, for orientation. The UWB range measurement is used in two
pipelines: first, it goes to the EKF; at the same time, it is fed to a Kalman
filter and then into the speed estimator. The speed estimation output goes
into the EKF (red line).  The robot heading is estimated by a complementary
filter~\cite{valenti_keeping_2015} that provides a quaternion estimation of
the orientation using an algebraic solution from the observation of gravity
and magnetic field. Finally, the range measurements, robot heading, and
speed estimation are fed into the EKF to estimate the robot pose.

Algorithm~\ref{alg1} outlines the approach in detail.  Sensor readings from
the IMU are fed into a complementary filter to obtain accurate heading in a
non-magnetic field inferred environment, as shown in line 10. The UWB range
and the computed heading are fed into an EKF to do a classical state
estimation for position, heading, linear and angular velocities (lines 11-12 in
Alg. \ref{alg1}).

The novelty of our system is the speed estimation, which corrects the velocity
estimation in the EKF. As explained above, the UWB range measurements are fed
into a separate Kalman filter to obtain smooth range measurements as $sRange$
(lines 13-14 in Alg.\ref{alg1}). When a change in $sRange$ is greater than the
threshold $TH_{distance}$ and the robot moves in a linear trajectory, the
speed estimator will calculate the speed. This speed is then used to correct
the velocity estimated from the EKF (lines 15-19). A variable window size
filter is applied to get a smooth speed estimate result (lines 6-7) as
discussed in \ref{sec:errorAna}.

\IncMargin{1em}
\begin{algorithm}\label{alg1}
	\small
	\SetAlgoLined
	\SetKwData{Left}{left}\SetKwData{This}{this}\SetKwData{Up}{up}
	\SetKwFunction{Union}{Union}\SetKwFunction{FindCompress}{FindCompress}
	\SetKwInOut{Input}{input}\SetKwInOut{Output}{output}
	\DontPrintSemicolon
	\SetKwFunction{FMain}{VelEstimator}
	\SetKwProg{Fn}{Function}{:}{}
	\BlankLine
	\Input{$time, range_t,  \boldsymbol{acc}_t,\boldsymbol{gyro}_t,\boldsymbol{magn}_t$ }
	\Output{$ \mathbf{x}_t $}
	
	$kRPs$ = ([range,timestamp],...) \# keyRangePairs\\
	\Fn{\FMain{$range, time$}}{

		newKeyPair = [$range, time$]\\
		$kRPs.append(newKeyPair)$\\
	 	$velList\leftarrow calculate\_velocity (kRPs_{i-2},kRPs_{i-1},kRPs_{i})$;\\
	 	$windowLen = range*ratio$\\
	 	$vel = average(velList[windowLen])$\\
	 	\KwRet vel \;
	}
	
	\While{True}{
		
		$heading_{t} = complementaryFilter(\boldsymbol{acc}_t,\boldsymbol{gyro}_t,\boldsymbol{magn}_t)$\\
		$\hat{\mathbf{x}}_{t} = \text{EKF}.predict()$\\
		$\mathbf{x}_{t} = \text{EKF}.update(range_t,heading_t)$\\
		$\hat{sRange} = \text{KF}.predict()$\\
		$sRange = \text{KF}.update(range_t)$\\
		\If{$|\boldsymbol{gyro}_t|<TH_{linear\_motion}$ and $(fdRange-range_i)>TH_{distance}$}{
			$i = i+1$\\
			$range_i$ = $sRange$\\
			$vel = VelEstimator(fdRange,time)$\\
			$\mathbf{x}_{t}.vel = w*\mathbf{x}_{t}.vel +(1-w)*vel$\\
			
			}
		
	}

\caption{State estimation algorithm from single UWB anchor.}
\end{algorithm}\DecMargin{1em}




\section{Experiments}
We validate our algorithm through simulations and real robot experiments. We
used two types of robots: a DJI M100 quadcopter and a
Duckiebot~\cite{paull_duckietown:_2017}. Robots are equipped with Decawave
DW1000\cite{noauthor_dw1000_nodate} based UWB modules \cite{ychiot}.
The quadcopter experiment specifically validated our speed estimation
algorithm and gave quantitative localization error based on GPS. The ground
mobile robot experiment shows that our algorithm can track very simple robots
even in complex trajectories. From the error summary in Table~\ref{table:ATE},
we can see our method has improved the accuracy by a factor 2 in simulations
and real robot experiments. Please note we use RMSE for simulation and
absolute trajectory error (ATE) for real robots experiments because we have
exact ground truth from simulations, but not for real robot experiments.

\begin{table}[h]
	\caption{Error comparison between EKF with or without speed estimator.}
	\begin{center}
		\begin{tabular}{|c||c|c|}
			\hline
		 Error (m) &Simulation (RMSE) & Drone Exp. (ATE)  \\
			\hline
			\hline
			\textbf{Without} speed estimator & 1.73 & 2.81\\
			\hline
			\textbf{With} speed estimator&  0.48&  1.05\\
			\hline
		\end{tabular}
	\end{center}
	\label{table:ATE}
\end{table}
	
\subsection{Simulation}

In our simulations, we generate a random trajectory with a differential wheel
robot kinematics model\cite{feng_where_1994}. There are five stages of motion, with different
headings and speed settings. The anchor is set at position (0,0) and the robot
starts at the point (10,0). The range measurement is corrupted by white Gaussian
noise with 0.2 m standard derivation, which is similar to the actual UWB
measurement derivation. The noise added to orientation is with a deviation of
0.1 rad.
	
	\begin{figure}[bp]
		
		\begin{center}
			\includegraphics[width=0.4\textwidth,center,trim=0 0.6cm 0 1.4cm, clip]{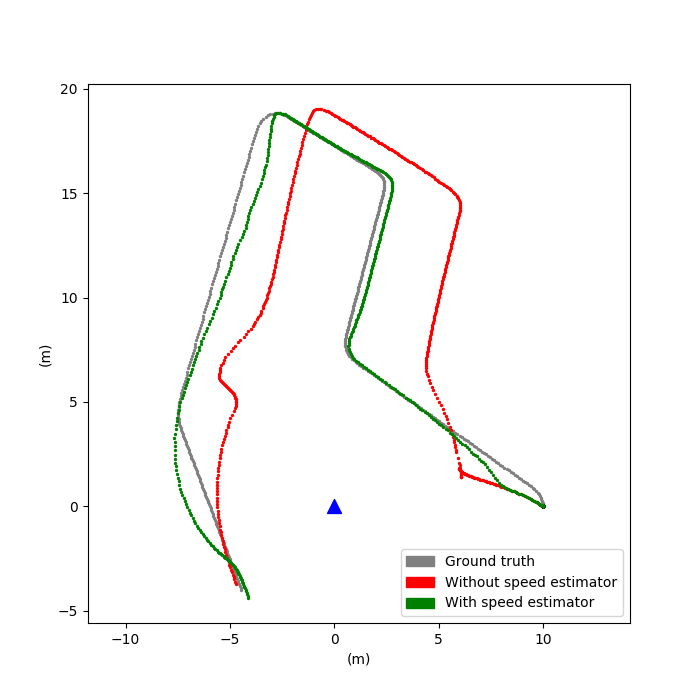}
			\caption{The trajectories of EKF with our speed estimator (proposed method, in green) and without speed estimator (vanilla EKF, in red), refer to the ground truth trajectory (gray).}
			\label{fig:simNoNoisyNoInitVelTraj} 
		\end{center}
	\end{figure}

\begin{figure}[!h]
		\subfloat{
			\hspace{-0.2cm}
			\includegraphics[trim={0 0 0 1.2cm}, width=0.5\textwidth,center,trim=0 0.6cm 0 1cm, clip]{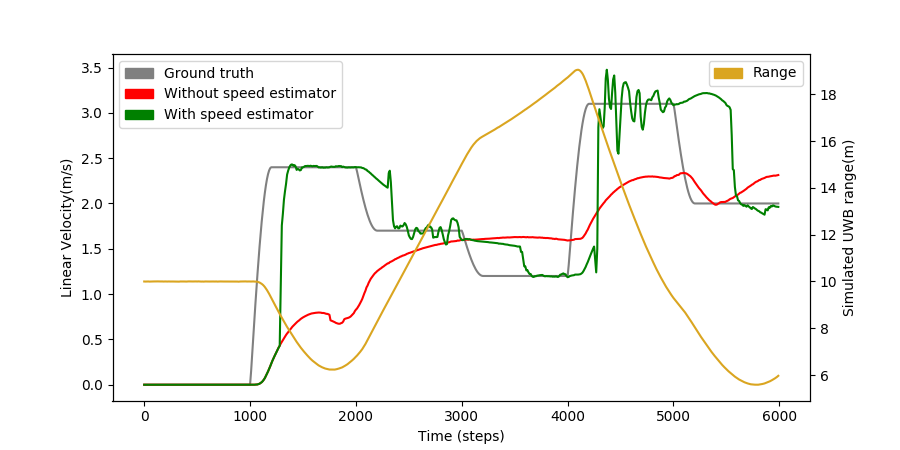}\label{fig:simWithNoisyInfo} }\\
		\subfloat{
			\includegraphics[trim={0 0 0 1.0cm},width=0.5\textwidth,center]{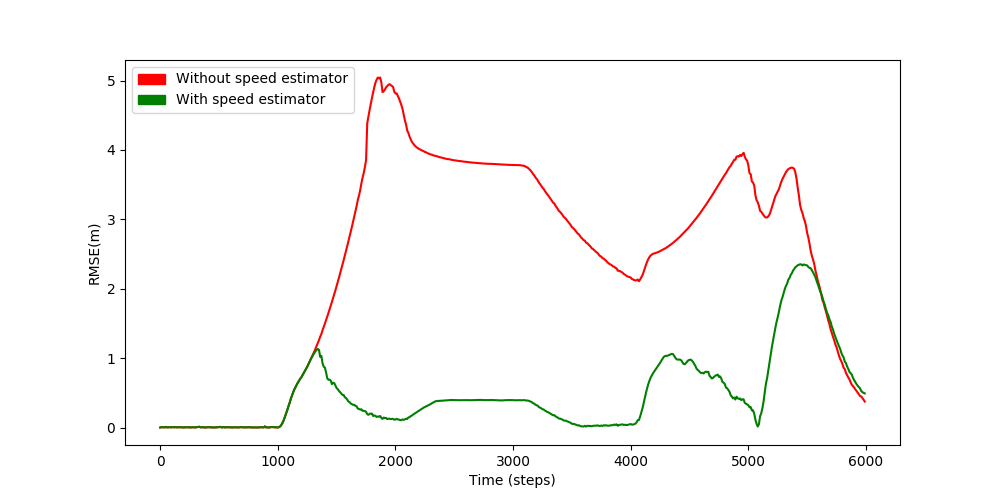}\label{fig:sim_rmse} }
		\caption{The top figure shows the velocity estimations of our method and vanilla EKF. We can see that although there are some delays, our algorithm can track the actual speed most times. However, the EKF without speed estimator can only catch up the trend of speed. The bottom figure is the comparison of RMSE. The error of our algorithm drops rapidly when the correct speed estimation is available.}
		\label{fig:simInfo}
		
	\end{figure}

As the top of Fig. \ref{fig:simInfo} shows, our method (green) can track the
ground truth velocity (gray) most of the time, which results in an accurate trajectory in Fig.\ref{fig:simNoNoisyNoInitVelTraj}. However, the vanilla EKF
model (red) cannot recover from the drift errors accumulated during the first
stage.  This proves that our algorithm can correct the accumulated error as
long as the speed estimator gives accurate estimations. More specifically,
compared with the bottom figure in  Fig. \ref{fig:simInfo}, the RMSE of our method drops rapidly (starts
from 1200 steps) when the speed estimation is available (around 1200 steps).
We reduce the RMSE of the vanilla EKF by more than 70\% and achieve the
accuracy of 0.48 m, which is impressive given the limited information.

\subsection{Speed Estimation for Flying Robot}

\begin{figure}[!h]
	\begin{center}
		\includegraphics[width=0.4\textwidth,center,trim=0 0.6cm 0 1.4cm, clip]{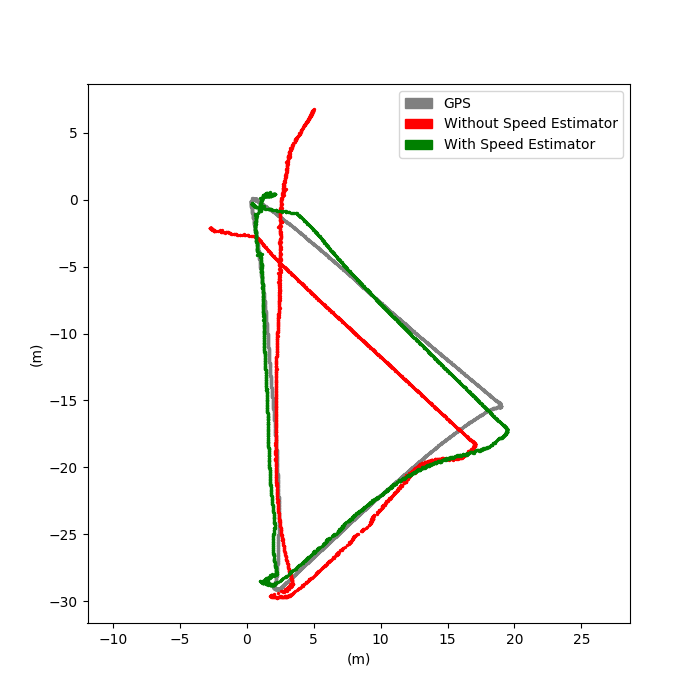}
		\caption{Experiments with a DJI M100 quadcopter. The robot is
                  programmed to fly in a triangle trajectory. The anchor is
                  placed on the ground. Our trajectory is much closer to the
                  GPS trajectory than the vanilla EKF without speed
                  estimator.}
		\label{fig:ATE} 
	\end{center}
\end{figure}

\begin{figure}[!h]
	\begin{center}
		\includegraphics[width=0.5\textwidth,center]{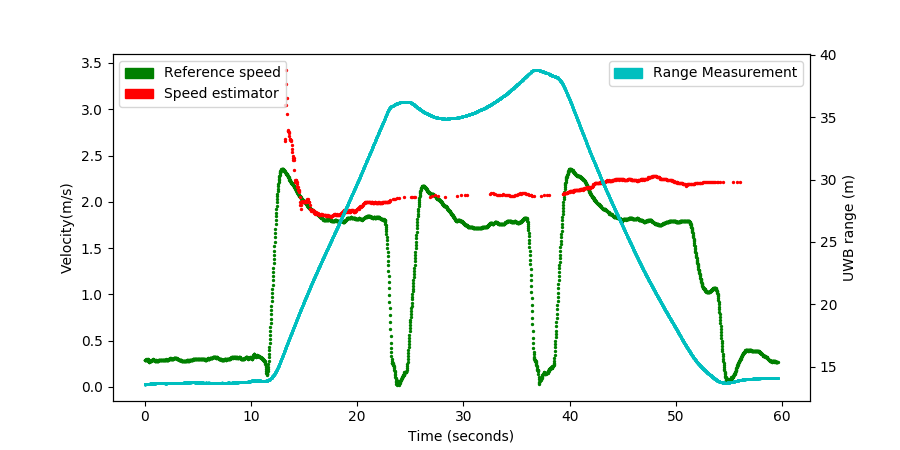}
		\caption{In the quadcopter experiment, we can estimate the speed
                  (red) simply by using the range measurement (cyan). The
                  reference from the DJI\_SDK is plotted in green.}
		\label{fig:m100vel} 
	\end{center}
\end{figure}

We have used a quadcopter (DJI M100) to validate our tracking and speed
estimation in the real world, which also illustrates the potential use for 3D
applications as well. We programmed a triangle trajectory with a speed
parameter of 2 m/s for the M100. First, we compare the estimated results with
the velocity feedback from DJI\_SDK software. Fig.\ref{fig:m100vel} shows our
algorithm can give correct speed estimation (red) based on range measurements
(cyan). Then we calculate ATE between our estimated trajectory and GPS
trajectory. As Fig.\ref{fig:ATE} and table \ref{table:ATE} show, the trajectory from
our algorithm is much closer than that of the EKF without our speed
estimator. With only one anchor setup, we get around 1 m ATE error, which is
much higher than the vanilla EKF with the accuracy of 2.8 m. Please also note
that the DJI velocity feedback in Fig.\ref{fig:m100vel} only works as
reference, instead of ground truth: for the first ten seconds, the robot is
hovering but the speed indicated from DJI\_SDK is around 0.3 m/s.

\subsection{Tracking of Ground Mobile Robot}
The ground robot platform is a simple indoor robot, the
Duckiebot~\cite{paull_duckietown:_2017}. Our version of the Duckiebot has an
RPi2 on-board computer. We add a 9DOF IMUs sensor and a UWB module to the
robot, as shown in Fig.\ref{fig:outdoorExpTraj}. The robot does not have wheel
encoders to get the speed or displacement. We manually control the robot to run a
university logo trajectory (TUM) in an outside basketball court.  We use our
localization system to track the robot pose. Fig.\ref{fig:outdoorExpTraj}
shows that the algorithm can track the trajectory correctly. This experiment
shows a qualitative evaluation, indicating that our algorithm can be easily
applied to simple robots and IoT devices.

\section{Conclusion and Discussion}
In this paper, we propose a localization algorithm for robots that are
equipped with low-cost IMUs and UWB sensors in an environment configured with
only a single UWB anchor. We estimate the speed from UWB range changes, which
makes the system temporally observable. Our algorithm effectively reduces the
accumulated errors by 60\%. With this algorithm, a large number of devices can
be localized, including IoT devices or cellphones with IMU and UWB sensors.



\bibliographystyle{IEEEtran}
\bibliography{IEEEabrv,singleAnchorRef}

\end{document}